\documentclass[a4paper,12pt,titlepage,oneside]{kumsthesis}

\usepackage{url}
\usepackage[bottom]{footmisc}
\usepackage[caption=false,font=footnotesize]{subfig}
\usepackage{amsmath}
\usepackage{amssymb}
\usepackage{amsfonts}
\usepackage{breakcites}

\usepackage{enumerate}
\usepackage[
  separate-uncertainty = true,
  multi-part-units = repeat
]{siunitx}
\usepackage{amsmath}
\usepackage{amssymb}

\graphicspath{{./Figures/}}

\usepackage{xspace}
\usepackage{epstopdf}

\newcommand \COMMENT  [1] {}       

\begin{document}

\Title{Utilizing coarse-grained data in low-data settings for event extraction} \Author{Osman Mutlu} \Year{February 15, 2022}
\TTitle{Olay bilgisi çıkarma sistemleri için az veri senaryosunda az detaylı veri kullanmak} \TYear{15 Şubat 2022}
\Program{Computer Science and Engineering}
\TProgram{Bilgisayar M\"{u}hendisli\u{g}i}
\Signature{Prof. Dr. Deniz Yuret (Advisor)}
\Signature{Asst. Prof. Dr. Reyyan Yeniterzi}
\Signature{Prof. Dr. Engin Erzin}

\prelimpages
\titlepage
\thesissignaturepage
\dedication{Dedicated to those who believe in me no matter what}
\abstract{
Annotating text data for event information extraction systems is hard, expensive, and error-prone. We investigate the feasibility of integrating coarse-grained data (document or sentence labels), which is far more feasible to obtain, instead of annotating more documents. We utilize a multi-task model with two auxiliary tasks, document and sentence binary classification, in addition to the main task of token classification. We perform a series of experiments with varying data regimes for the aforementioned integration. Results show that while introducing extra coarse-grained data offers greater improvement and robustness, a gain is still possible with only the addition of negative documents that have no information on any event.
}
\oz{Olay bilgisi çıkarma sistemleri için metin verisi işaretlemesi yapmak hem zor, hem pahalı, hem de hata yapmaya oldukça açıktır. Bu tezde, yeni detaylı işaretleme yapmak yerine, çok daha kolay şekilde elde edilebilen daha az detaylı (döküman ve cümle etiketlemesi) veri kullanmanın fizibilitesini ölçüyoruz. Döküman ve cümle etiketlerini kullanmak için çok amaçlı modelimizi, ana işimiz olan kelime sınıflandırmasının yanında döküman ve cümle ikili sınıflandırması yan işleri ile eğitiyoruz. Bu amaçta, değişen veri rejimleri içeren birtakım deneyler icra ediyoruz. Deneylerin sonuçları bu eklenen daha az detaylı verinin daha iyi performans ve stabiliteye yol açtığını gösterirken, aynı zamanda orijinal veriye sadece içinde hiçbir şekilde olay bilgisi bulundurmayan negatif dökümanlar eklemenin göz ardı edilemeyecek katkısını da gözler önüne seriyor.}
\acknowledgments{
I would like to express my gratitude to my advisor, Prof. Dr. Deniz Yuret for his support and vast knowledge. I would also like to thank Prof. Dr. Engin Erzin and Asst. Prof. Dr. Reyyan Yeniterzi for their participation in my thesis committee.

I would like to thank my fellow members of AI Lab, Ilker Kesen, Ozan Arkan Can, Omer Kirnap, Ali Safaya, Ulas Sert, Cemil Cengiz for their acceptance, support and companionship.

I acknowledge the funding of my master's studies by the European Research Council (ERC) Starting Grant 714868 awarded to Dr. Erdem Y\"{o}r\"{u}k for his project Emerging Welfare.

Finally, I am forever grateful to my co-advisor Dr. Ali Hurriyetoglu for his continuous support, guidance, motivation and trust. This thesis would not be possible without him.
}
\tableofcontents
\listoftables
\listoffigures
\abbreviations{
\begin{tabular}{lp{1cm}l}
    NLP & & Natural Language Processing \\
	DL & & Deep Learning \\
	BERT & & Bidirectional Encoder Representations from Transformers \\
	GRU & & Gated Recurrent Unit \\
	MLP & & Multi-Layer Perceptron \\
	MLM & & Masked Language Modeling \\
	NSP & & Next Sentence Prediction
\end{tabular}
}
\textpages

\chapter{Introduction}
\label{intro}


The recent surge of development in Natural Language Processing (NLP) can be attributed to the ever-increasing power of the hardware, advancements in architecture design of Deep Learning (DL) models, and most importantly data size. But acquiring data is not always straightforward, especially for NLP tasks such as event information extraction. Annotating documents for event information extraction is a strenuous task that requires domain expertise in both creating the specifications and the actual act of annotation. These specifications often take time to develop throughout the annotation process, even requiring documents to be annotated retroactively. Due to the difficult and time-consuming nature of this process and the usual requirement of expertise, it ultimately becomes quite an expensive venture to undertake. Moreover, this inherent difficulty affects the resulting annotation quality, model performance, and the reliable estimation of task performance. This effect is often observed through inter-annotator agreement scores. This interplay between the size and the quality of the data affects what automatic approaches can achieve.

The goal of event extraction is to detect events, which are occurrences or a change of state with a time and a place, and its arguments, which are participant(s) or attributes of an event, in a given text.
We use the definitions of ACE\footnote{See https://www.ldc.upenn.edu/sites/www.ldc.upenn.edu/files/english-events-guidelines-v5.4.3.pdf for a description of this task.} to discuss about the characteristics of an event:
\begin{itemize}
    \item Event trigger: the main word or span of words that most clearly expresses an event occurrence, typically a verb/verb phrase or a noun/noun phrase.
    \item Event argument: an entity mention, temporal expression or value that serves as a participant or attribute with a specific role in an event.
\end{itemize}
An example event with its trigger and arguments can be seen in Figure \ref{example_human_readable}. Event extraction problem exists in several domains; it is used to extract adverse drug effects from biomedical text \cite{liu2015identifying,wei2020study}, in legal applications \cite{shen-etal-2020-hierarchical}, to extract financial information \cite{dor2019financial} from news, and in several other domains.
We consider event extraction as a token classification task as it is commonly deemed by DL approaches. This is achieved by using BIO format \cite{ramshaw-marcus-95-text} in the output where each token (word, sub-word or punctuation mark) is assigned a label. In BIO format, for each type (be it event trigger or argument) that needs to be extracted, there are two labels starting with ``B-'' and ``I-'', indicating the beginning token and intermediate token of the span of tokens. All the tokens that do not belong to any span to be extracted are labeled as ``O'', as in other. The conversion of the previous example into BIO format can be observed in Figure \ref{example_bio}. So, for a token classification task, the inputs are the words of a document and the outputs are the corresponding labels for each word.

There have been many efforts for alleviating the issues caused by limited gold standard data. Data-driven and model-driven approaches are the prominent proposals to tackle this problem. Data-driven approaches include; using pseudo labels generated by previously trained models~\cite{elaraby-litman-2021-self}, applying data augmentation ~\cite{Liu-et-al-2020-a-survey,chen2021description,wang-et-al-2016-a-multiple-instance}, utilizing distant supervision~\cite{mintz-etal-2009-distant} and knowledge base creation~\cite{meng-etal-cnn-iets-a-cnn} techniques. However, these approaches require some initial gold-standard data, to train a model and to augment. Sometimes, they even achieve worse results than of a baseline trained with said initial gold-standard data. Model-driven approaches consist mainly of pre-training\cite{caselli-etal-2021-protest} and zero-shot~\cite{lyu-etal-2021-zero} learning methods. The main solutions introduced by zero-shot models are data selection and transfer learning techniques. Unfortunately, these models are still in the developmental stage and demonstrate subpar performances to even data-driven approaches. 

Since the annotation process of event extraction (or in short, token annotation) is inefficient at best and damaging at worst, we investigate the possibility of employing coarser data containing similar information, which would be more reliable and easier to get, in line with recent proposals for benefiting from coarse-grained data for extracting fine-grained information\cite{rei-sogaard-2018-zero,zhao-etal-2018-document,dale-etal-2021-skoltechnlp}. The aforementioned coarser data are document or sentence binary labels that correspond to whether or not that instance contains information on an event. Labeling of such data is relatively effortless, effectively circumventing most of the aforementioned issues with the annotation of event information extraction documents. Since labeling document or sentences require more relaxed specifications and less linguistic expertise than token annotations as the semantic roles contribute to the annotation process, it is considerably cheaper and easier to achieve a higher quality of data.
To utilize this coarse data, we introduce two new auxiliary tasks in addition to the main token classification task, namely document and sentence binary classification tasks and train a hierarchically structured multi-task model.
In our experiments, we analyze the effect of the addition of the two auxiliary tasks to our training process, with extra document or sentence binary labels. Our results show that adding extra coarse-grained data offers improvement to the main task of token classification and robustness specifically in very low data sizes.

The structure of this thesis is organized as follows:

\textbf{Chapter \ref{relwork}} expands on the relative work.

\textbf{Chapter \ref{data}} presents the data set utilized in this thesis and describes the main and auxiliary tasks.

\textbf{Chapter \ref{model}} describes the model structure used for the experiments.

\textbf{Chapter \ref{chapter:experiments}} illustrates information about the experiments and their results.

\textbf{Chapter \ref{conclusion}} concludes the work presented in this thesis and discusses possible future work.

\textbf{Appendix} displays the detailed results for all the experiments.

\chapter{Related Work}
\label{relwork}

Allaying the hardships of text data annotation has been the focus of plenty of studies. The main goal of such an effort is to be able to maintain the performance when decreasing the amount of annotation. We divide the previous methods into two categories: data-driven and model-driven approaches.

\section{Data-driven Approaches}
Using silver quality data, which is of lower quality than the manually annotated or labeled gold standard data, is the single-most common technique to alleviate the low data problem. Data augmentation, self training and distant/weak supervision, where new examples are generated (generation phase) employing whichever technique and then used in training a new model (training phase) for the actual task in hand, can all be collected under a single umbrella definition of data-driven approaches.

\subsection{Data Augmentation}
Data augmentation is a method used to automatically populate the available data with slight variations utilizing different techniques. It was first commonly used in the field of computer vision, achieved by rotating, translating, adding noise and compressing, and later introduced to the NLP field by means of back translation and lexical substitution. Back translation is realized with two translations, one from the original language of the data to a target language, and the other from the target language back to the original language. This process often creates different word ordering or variations on the original data due to differences in languages. Lexical substitution is a relatively simple technique, where some percentage of the words are replaced with their synonym or syntactically correct substitutes (a noun is replaced for another noun for example). \cite{Liu-et-al-2020-a-survey} provides a detailed summary of the previous work.

\subsection{Self training}
Self training~\cite{yarowsky1995unsupervised} is the process of using previously annotated examples (often relatively small in size) to train a model that then labels ample amount of unlabeled data to be used in training of the final model. After this pseudo labeling process, newly acquired data is mixed with the original data to train a final model. \cite{elaraby-litman-2021-self} uses the generated pseudo labels to train their model before fine-tuning with gold examples.

\subsection{Distant/Weak supervision}
The objective of distant/weak supervision is to create pseudo labels from large unlabeled examples using a related task's data to the original task. These pseudo labels are later used in training for the final model. For example; a model trained with a question answering data set can be used to automatically identify the arguments of events by prompting the model with ``What'', ``Who'', ``Where'', ``When'' questions. To train their model, both \cite{mintz-etal-2009-distant,chen2017automatically} automatically generate labeled examples by employing a semantic knowledge base. \cite{huang2012bootstrapped} generate their labels utilizing patterns with manually created noun lists for every event type.

\section{Model-driven Approaches}
Model-driven approaches include zero-shot learning and pre-training with relative data, where the approaches change the model in some way.

\subsection{Zero-shot learning}
Per their definition, zero-shot learning approaches do not use any training data for the task at hand. They are similar to distant/weak supervision techniques in that both transfer knowledge from related tasks. However, contrary to distant/weak supervision, there is no training process with gold or silver data for the original task; zero-shot learning methods have a single prediction phase instead of two previously mentioned generation and training phases. \cite{lyu-etal-2021-zero} extracts the triggers by a combination of textual entailment and semantic role labeling models, and predicts its event arguments using a pre-trained question answering model. \cite{rei-sogaard-2018-zero} trains a model with soft attention utilizing sentence level labels, later inferring the token labels of the test data with said attention mechanism.

\subsection{Further Pre-training}
In recent years, pre-trained language models such as BERT~\cite{Devlin+19} and ELMO~\cite{peters2018deep} gained extensive popularity and they managed to boost many NLP tasks' performances. Researchers also started to further pre-train these models with self-supervised examples from the intended task's domain to achieve higher scores. This further pre-training is also applied to the low-data problem by further pre-training with data from supervised tasks related to the original task. \cite{dale-etal-2021-skoltechnlp} uses sentence labels to further pre-train their model before fine-tuning with token annotations. Both \cite{zhangcan,caselli-etal-2021-protest} pre-trains their model with unlabeled examples from the domain of their original task to achieve better performances.

\section*{}
Our approach is similar to further pre-training in that we also utilize a relevant task's data (sentence and document binary classification data) that can be used in training the same model, but we actually train both tasks at the same time instead of in order. We achieve this by implementing a multi-task model with a hierarchical architecture inspired by ScopeIt~\cite{patra-etal-2020-scopeit}.

\chapter{Data and Tasks}
\label{data}
In this chapter, we describe the data set we employed, and our main and two auxiliary tasks.

\section{Data Set}
We utilize ACL CASE 2021 shared task~\cite{hurriyetouglu2021multilingual} data, which is an event information detection shared task focusing on protest events. This data set consists of news articles from various countries published between 2000 and 2018 in various languages. 
The shared task is divided into 4 subsequent tasks consisting of data with different granularities to better simulate a real-world event extraction pipeline. We particularly use a subset of the English training data from subtask 4, which is the token classification task, and the same English test data for our experiments. 

\begin{figure}[!h]
\begin{center}
\label{example_human_readable}
\underline{In August that year}, militantly aggressive \underline{Kurmis} \textbf{mowed down} 14 \underline{Dalits} in \underline{Nalanda district}.
\caption{An event extraction annotation example. Bold words represent the trigger of the event, and underlined words represent various arguments of the event.}
\label{example_human_readable}
\end{center}
\end{figure}

There are a total of 717 annotated documents for training. These annotations were distributed in BIO format, meaning there are no overlapping labels for any individual token, effectively turning this task into token classification. A sample in BIO format can be seen in Figure \ref{example_bio}. A human readable format of the same example is shown in Figure \ref{example_human_readable}, the bold face indicating the event trigger and the underlined tokens specifying the arguments.

\begin{figure}[!h]
\begin{center}
In\textsubscript{B-etime} August\textsubscript{I-etime} that\textsubscript{I-etime} year\textsubscript{I-etime} ,\textsubscript{O} militantly\textsubscript{O} aggressive\textsubscript{O} Kurmis\textsubscript{B-participant} mowed\textsubscript{B-trigger} down\textsubscript{I-trigger} 14\textsubscript{O} Dalits\textsubscript{B-target} in\textsubscript{O} Nalanda\textsubscript{B-place} district\textsubscript{I-place} .\textsubscript{O}
\caption{BIO equivalent of example in Figure \ref{example_human_readable}}
\label{example_bio}
\end{center}
\end{figure}

The test set, which is the same throughout the experiments since we only care about token classification task's scores, includes 179 annotated documents. There are no test sets for document or sentence level classification tasks. We present the distribution of annotations for training and test sets in Table~\ref{tab1}.

\begin{table*}[ht]
\begin{center}
\caption{The number of each annotation for training and test data sets.}
\begin{tabular}{|c|c|c|c|c|c|c|c|}
\hline
\multicolumn{1}{|l|}{} & \textbf{etime} & \textbf{fname} & \textbf{organizer} & \textbf{participant} & \textbf{place} & \textbf{target} & \textbf{trigger} \\ \hline
\textbf{Train}         & 1071           & 1089           & 1187               & 2435                 & 1436           & 1334            & 4096             \\ \hline
\textbf{Test}          & 260            & 224            & 223                & 542                  & 313            & 286             & 929              \\ \hline
\end{tabular}
\label{tab1}
\end{center}
\end{table*}

\section{Main task}
As mentioned before, our main task is a multi-class token classification task. The objective of this task is to assign a label to each given word.
In our experiments, we use the data set mentioned above, 717 documents for training and 179 documents for testing. There are a total of 15 labels including the ``O'' label, and two labels representing ``B-'' and ``I-'' for each annotation type in Table~\ref{tab1} to classify into. To evaluate our experiments, we use a Python implementation\footnote{\url{https://github.com/sighsmile/conlleval}, accessed on June 6, 2021.} of the original conlleval evaluation script\footnote{\url{www.cnts.ua.ac.be/conll2000/chunking/conlleval.txt}, accessed on June 11, 2021.}, which we simply refer as F1 score.

\section{Auxiliary Tasks}
We utilize two auxiliary tasks to help in training, document and sentence binary classification tasks.

\subsection{Document binary classification}
One of the auxiliary tasks is document binary classification. The objective of this task is to assign ``negative'' or ``positive'' to a given document. A document is ``positive'' if it contains any event information and ``negative'' otherwise. As mentioned above, our subset of the shared task data includes 717 documents with annotations. We consider these 717 documents as “positive” since they contain events. In certain experiments, we supplement this data with negatively labeled documents to maintain label balance. These “negative” documents were obtained from the same shared task’s subtask 1 for document classification. This is the only additional data we used throughout our experiments.

\subsection{Sentence binary classification}
The other auxiliary task is sentence binary classification. The objective of this task is to assign ``negative'' or ``positive'' to a given sentence. A sentence is ``positive'' if it contains any event information and ``negative'' otherwise. On average, there are 14.06 sentences in the annotated 717 documents. 29\% of these sentences are positively labeled as they contain at least one token annotation. We label the rest of the sentences as negative, totaling 2,893 positive and 7,191 negative sentences. There are no additional data used for sentence classification task in any experiment other than the available 717 annotated documents.

\chapter{Model Structure}
\label{model}

In this chapter, we outline the model structure used in all the experiments. Starting from the representations of the model's input and outputs, we delve into each neural module and finally describe how these components come together aided with a couple of figures.

The model in question has a multi-task architecture inspired by ScopeIt~\cite{patra-etal-2020-scopeit}. Our implementation can be found in github\footnote{https://github.com/OsmanMutlu/ms\_thesis}.

\begin{figure*}[ht]
\begin{center}
\includegraphics[scale=0.5]{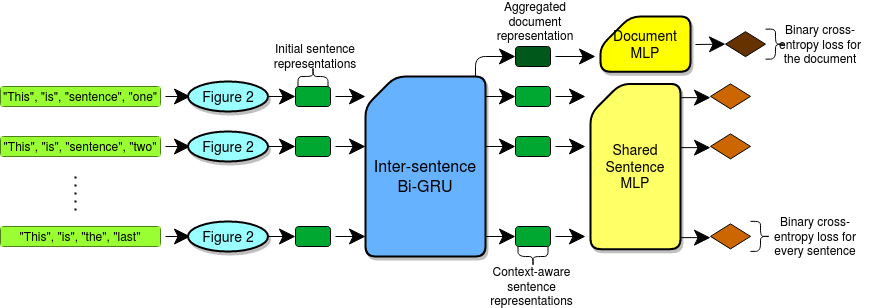} 

\caption{The main structure of our multi-task model. ``Figure 2'' refers to the process each sentence goes through, which can be seen in Figure \ref{fig2}.}
\label{fig1}
\end{center}
\end{figure*}

\section{Input and Output Representations}
Input to the model is a document with $S$ sentences each containing $N$ words. In our experiments, we cut or pad each sentence to 128 tokens in order to process them in parallel. We also cut the documents at 200 sentences. These numbers were selected with distribution of lengths of sentences and GPU size in mind. Each sentence is input separately to a shared instance of BERT encoder.

There are three outputs from the model, representing each task; document classification, sentence classification and token classification. A single label of positive or negative is produced for document classification task. For sentence classification task, $S$ positive or negative labels are assigned for each sentence respectively. Finally, $N$ labels are assigned for each $S$ sentence, totalling up to $N\times S$ labels.

\section{Neural Modules}
In this section, we first specify each module of the model to better communicate how they are unified to form the whole structure.

\subsection{Encoder}
The encoder of the model consists of two parts: a BERT encoder and a bidirectional Gated Recurrent Unit (GRU)~\cite{cho2014properties}.

\subsubsection{BERT Encoder}
\begin{figure*}[ht]
\begin{center}
\includegraphics[scale=0.5]{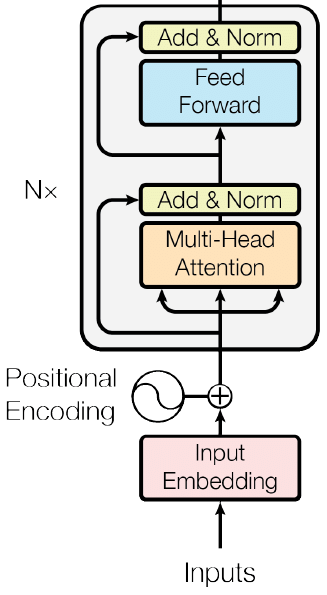} 

\caption{A simplification of the transformer encoder from~\cite{vaswani2017attention}}
\label{encoder_fig}
\end{center}
\end{figure*}

BERT encoder is a pre-trained version of the original Transformer~\cite{vaswani2017attention} encoder-decoder, specifically the encoder part, which was used in machine translation. It was trained on English Wikipedia and BooksCorpus~\cite{zhu2015aligning} using a couple of self-supervised tasks with the purpose of general natural language understanding in mind. 
The tasks in question are Masked Language Modeling (MLM) and Next Sentence Prediction (NSP). MLM is a variant of standard language modeling objective devised by the authors with Transformer architecture in mind. Since Transformer encoder processes all the input words in parallel, meaning standard language modeling objective of predicting the next word given the previous words fail, the authors morph the objective into correctly predicting 15\% of the input words which are masked beforehand. The objective of NSP task is to correctly predict if two input sentences were sequential or not in the original text.
We use the freely distributed weights\footnote{https://github.com/google-research/bert} (specifically BERT-base-uncased) of the BERT encoder as our starting weights. As shown in the original paper, this helps the model extremely even in low-resource tasks. An overview of the encoder can be seen in~\ref{encoder_fig}.

All $[w_{i1}, ..., w_{iN}]$ words in a sentence is passed through a shared instance of BERT encoder, and it outputs $[\Vec{b_{i1}}, ..., \Vec{b_{iN}}]$ vectors of hidden size length.

\subsubsection{Intra-sentence Bi-GRU}
The first bi-GRU, called intra-sentence bi-GRU, aggregates $[\Vec{b_{i1}}, ..., \Vec{b_{iN}}]$ vectors representing the words in a sentence to form a single vector $\Vec{s_i}$ for each sentence. This is achieved by adding the final hidden representations of the bi-GRU together.


\begin{align}
    \Vec{c_{ijb}} = \overleftarrow{GRU_{b}}(\Vec{b_{ij}}) \label{eq:1} \\
    \Vec{c_{ijf}} = \overrightarrow{GRU_{f}}(\Vec{b_{ij}}) \label{eq:2} \\
    \Vec{c_{ij}} = \Vec{c_{ijb}};\Vec{c_{ijf}} \label{eq:3} \\
    \Vec{s_{i}} = \Vec{c_{i1b}} + \Vec{c_{iNf}} \label{eq:4}
\end{align}

The $[\Vec{b_{i1}}, ..., \Vec{b_{iN}}]$ vectors that the BERT encoder outputs are input to intra-sentence bi-GRU, and it outputs $[\Vec{c_{i1}}, ..., \Vec{c_{iN}}]$ vectors for each word and a single $\Vec{s_i}$ vector for the sentence. These $[\Vec{c_{i1}}, ..., \Vec{c_{iN}}]$ vectors are the final representations for the words to be fed into the first (token) Multi-layer Perceptron (MLP).

\subsection{Inter-sentence Bi-GRU}
A second bi-GRU, dubbed inter-sentence bi-GRU, is used to aggregate the vectors of each sentence of the document. This aggregation enables the model to generate a single vector to represent the document, and it enables each sentence's representation to ``know" about neighbouring sentences, or to become context-aware.

\begin{align}
    \Vec{p_{ib}} = \overleftarrow{GRU_{b}}(\Vec{s_{i}}) \label{eq:5} \\
    \Vec{p_{if}} = \overrightarrow{GRU_{f}}(\Vec{s_{i}}) \label{eq:6} \\
    \Vec{p_{i}} = \Vec{p_{ib}};\Vec{p_{if}} \label{eq:7} \\
    \Vec{d} = \Vec{p_{1b}} + \Vec{p_{Sf}} \label{eq:8}
\end{align}

Inter-sentence bi-GRU is fed with $[\Vec{s_1}, ..., \Vec{s_S}]$ sentence vectors from the previous bi-GRU, and it outputs $[\Vec{p_1}, ..., \Vec{p_S}]$ context-aware vectors and a single $\Vec{d}$ vector for the document generated by adding the final hidden representations of the bi-GRU together. The $[\Vec{p_1}, ..., \Vec{p_S}]$ context-aware vectors are the final representation for the sentences to be fed into the second (sentence) MLP.

\subsection{MLP's}
The model has three MLP's with two layers for the three classification tasks; token, sentence and document classification. The final representations of each task is input to its corresponding MLP and a probability distribution of labels is received for loss calculation.

\section{Outline}
\begin{figure}[!h]
\begin{center}
\includegraphics[scale=0.70]{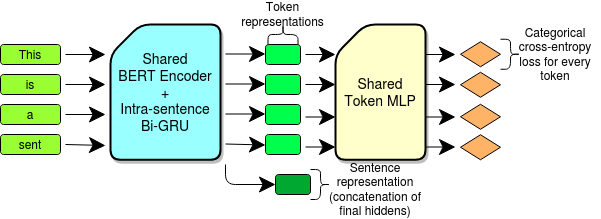} 

\caption{The process each sentence goes through.}
\label{fig2}
\end{center}
\end{figure}

We combine three cross-entropy losses calculated from three levels that are document, sentence, and token level. For the token level, we process each sentence of a document separately with the BERT encoder and the first bi-GRU, which is dubbed intra-sentence bi-GRU, on top of it. We then pass each token embedding that is generated by this step through the first MLP. The categorical cross-entropy loss is calculated over these tokens. This process can be seen in Figure \ref{fig2}. For sentence level, we take each aggregated sentence embedding (adding each direction's last hidden together from the intra-sentence bi-GRU) and pass them through a second bi-GRU, which is dubbed inter-sentence bi-GRU. The resulting sentence representations can be seen as context-aware due to the nature of the bi-GRU. We again pass these representations through a second MLP and calculate binary cross-entropy loss. For the document level, we add each direction's last hidden together from the inter-sentence bi-GRU and pass it through a third MLP to calculate binary cross-entropy loss. After scaling each loss, we add all together to get a final loss value.


\chapter{Experiments and Results}
\label{chapter:experiments}
In this chapter, we describe our experiments, and share their results and the details of the training process.

\section{Experiment Descriptions}
\label{expdesc}

We deploy three main experiment sets associated with three main research questions, each consecutive experiment set adding new data; first using the inherent sentence and document binary classification data in the annotated documents, second adding easily available negative documents to the mix, and third adding never before seen positive document and sentence binary classification data. In each experiment set, there are three experiments based on the combination of losses we train with in addition to token classification loss; only sentence classification loss (variation 1), only document classification loss (variation 2), and both sentence and document classification losses (variation 3). This ensures that we pinpoint the effect of each individual auxiliary task. All of our experiments are run 3 times to calculate the average performance and standard deviation of the performance scores. We also run all the experiments while gradually decreasing the amount of the data to measure the effect of the data size in these configurations. This gradual decrease also allows us to use the discarded examples as the extra (never before seen) document and sentence classification data in experiment set 3.

\subsubsection{Baseline:}
We use the 717 annotated documents to train our model as our baseline with only main token classification task's loss. This baseline does not change throughout our results. Even though some weights of the model do not update due to the architecture in some cases, we always train the same model for all our experiments. For our baseline model, the inter-sentence bi-GRU, sentence and document MLP's do not train since we only use cross-entropy loss for each token, but we technically use the same model. This is important for a fair comparison. 

\subsubsection{Experiment Set 1:}
Although we talked about the relative ease and reliability of labeling document and sentence binary data, for the first experiment set no extra data is used outside of the available annotated documents. We first investigate the use of the inherent information already available in token annotations. This inherent information refers to sentence and document labels aforementioned in Chapter~\ref{data}. This setting allows us to measure the effect of introducing our auxiliary tasks to the baseline without any change of data. This is important because of two reasons. Firstly, it enables us to set a reference point for the loss variations in the other two experiment sets. Secondly, it enables us to determine whether the fine-grained task of token classification actually encompasses the auxiliary tasks when training.

\subsubsection{Experiment Set 2:}
Introducing the document classification task to the baseline in our first experiment set has a major flaw since all of 717 annotated documents are positive. The absence of any negative documents severely affects the training process, and it hinders any positive effect that employing document classification loss may have. So, we include negative documents to balance out the positive ones, obtained from another subtask of the same shared task as mentioned in Chapter~\ref{data}. For example; if we use 501 token annotated documents whilst training, we add 501 negatively labeled documents. Since the only change occurs when we calculate document classification loss, there is no need to repeat this experiment for loss variation 1 (only sentence classification loss).

\subsubsection{Experiment Set 3:}
Finally, we experiment with the effects of including extra data for the auxiliary tasks. In order to simulate a real-world situation where one decides how many documents to annotate, we change our previous scenario of gradually decreasing data size slightly. Instead of completely discarding a certain percentage of the data, we use that percentage of documents as extra data to train with only sentence and document classification tasks. To clarify, imagine the following scenario; a researcher labeled 717 documents as positive -containing event information- and the sentences of these 717 documents as negative and positive. After the painstaking process of creating a refined specification for the token annotations, the researcher realizes that they spent most of their budget which is not a surprise due to the iterative process of creating such a specification. So, they allocate their resources and time to annotate only a certain percentage of these 717 documents instead of all of them. The question is, how many of these documents should the researcher annotate without a significant decrease in token classification performance in relation to annotating all of them?

\section{Results}
\label{results}

Results are visualized using bar graphs for better understanding, but the detailed results can also be seen in appendices. The y axis is F1 score for token classification and x axis is the number of token annotated documents. Each color represents the task variations used in training; blue for baseline (only the main token classification task), green for sentence classification task in addition to token classification (variation 1), red for document classification task in addition to token classification (variation 2), and yellow for both sentence and document classification auxiliary tasks in addition to token classification. The black line in each bar indicates standard deviation.

\subsubsection{Experiment Set 1:}

The results from the first experiment set are close to our baseline; the slight variations can be attributed to the standard deviation of the three runs. We can easily conclude that there is no improvement without any introduction of new data to our model even though we start to utilize document and sentence classification tasks. Evidently, when training for the token classification task, the internal representations of our model already capture the required information for the auxiliary tasks. 

\begin{figure}[!h]
\begin{center}
\includegraphics{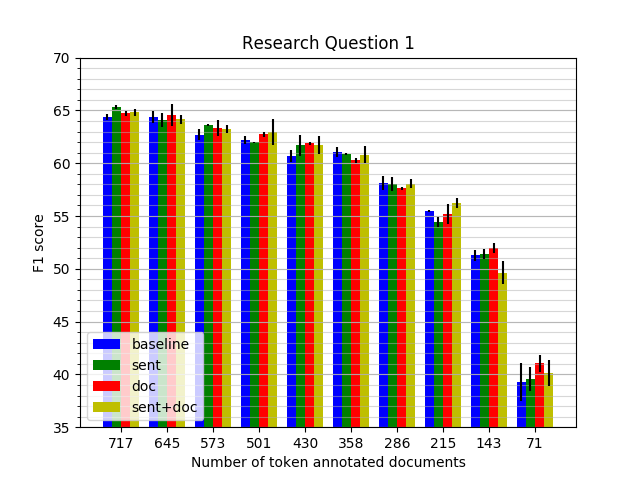} 

\caption{Results from experiment set 1, which focuses on the effects of our auxiliary tasks without any data addition. No improvement upon baseline is realized which signals that token classification task encapsulates the newly introduced auxiliary tasks.}
\label{fig_rq1}
\end{center}
\end{figure}

\subsubsection{Experiment Set 2:}

\begin{figure}[!h]
\begin{center}
\includegraphics{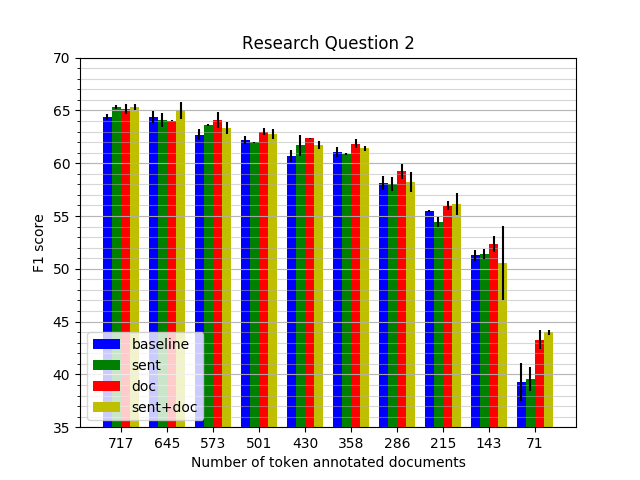} 

\caption{Results from experiment set 2, which focuses on the effect of adding negatively labeled documents with no event information. Even though there is no information related to any event in the added documents since they are negative, there is a performance increase.}
\label{fig_rq2}
\end{center}
\end{figure}

As can be seen from Figure~\ref{fig_rq2}, there is a definite improvement to the results under 80\% data size. We can surmise that supplementing negative documents to balance out the positive ones increases the performance of the model. Since negative documents are especially easy to acquire -they are a natural byproduct of the selection process of documents to be annotated at token level- this method can be used for a quick performance increase for current event extraction models. These results represent the first significant outcome of our experiments; even though there is no information related to any event in the documents we add, we can still gain improvement.

\subsubsection{Experiment Set 3:}
\label{exp_set3}

\begin{figure}[!h]
\begin{center}
\includegraphics{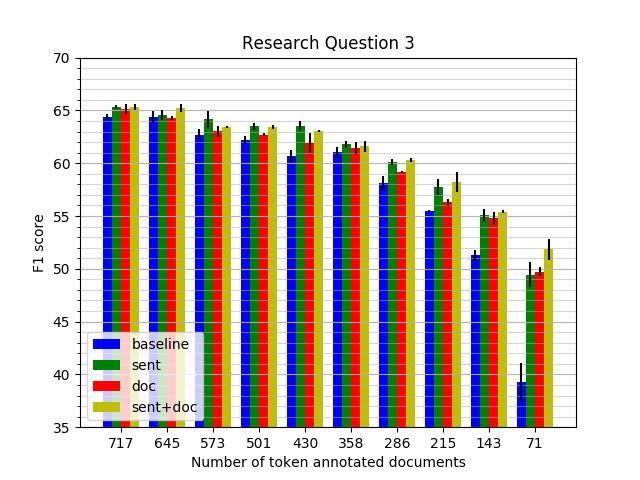} 

\caption{Results from experiment set 3, which focuses on the effects of adding extra coarse-grained data. There is a significant improvement to be gained when extra document or sentence data is introduced.}
\label{fig_rq3}
\end{center}
\end{figure}

\begin{figure}[!h]
\begin{center}
\includegraphics{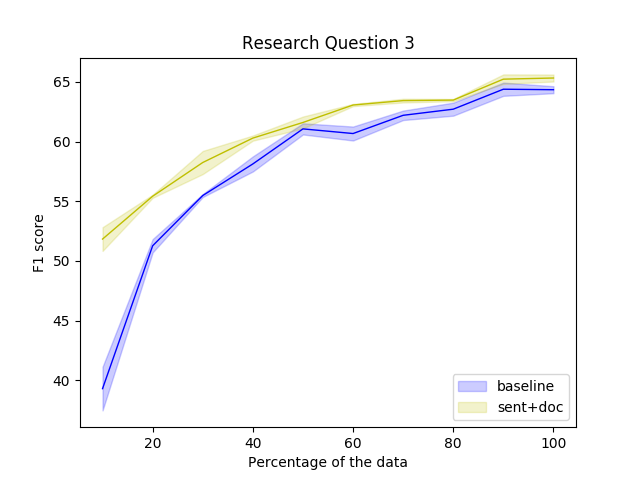} 

\caption{Partial results from experiment set 3, , which focuses on the effects of adding extra coarse-grained data. Only the results from baseline and the model with both auxiliary tasks is shown in a more pronounced way in this figure.}
\label{fig_line_rq3}
\end{center}
\end{figure}

We see a significant gain overall from Figure~\ref{fig_rq3}. Analyzing the results, for this instance, our question as to how many documents should a researcher annotate can be answered as 60\% of the original documents. 
Figure~\ref{fig_line_rq3} displays the results in a more pronounced way. As the number of token annotated documents decreases, the improvement that our auxiliary tasks and extra coarse-grained data provides is increased. When training with all the 717 token annotated documents, even though there is no extra positive documents or any extra sentence labels, our model performs better than baseline due to the addition of negatively labeled documents.
This experiment set causes two other sets to develop. Firstly, this experiment set has two changing variables; token data size and extra (not-seen in token classification task) document and sentence data size. We rectify this in experiment set 3.1 and set the extra data size to a fixed value. Secondly, there is a huge improvement in the smallest data size which requires further investigation in experiment set 3.2.

\subsubsection{Experiment Set 3.1:}

\begin{figure}[!h]
\begin{center}
\includegraphics{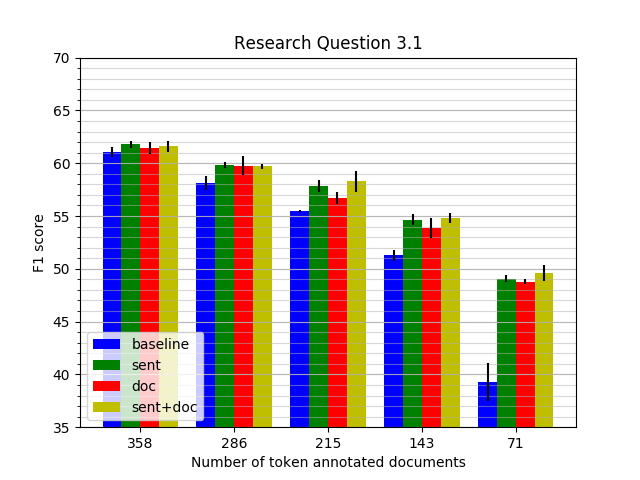} 

\caption{Results from experiment set 3.1, which is variation of experiment set 3 where extra data size is fixed. The results are similar to those of experiment set 3.}
\label{fig_rq3_fix_extra}
\end{center}
\end{figure}

We start from 50\% of the data and fix the discarded 50\% as extra coarse-grained data and use that part in all the other runs. For example, when using 10\% of the data in token classification, we do not use the extra discarded 40\% in auxiliary tasks. This enables us to observe the change in performance between experiments without confusion as to whether it originated from change in extra data size or token data size. The results support the original experiment set 3 in that the performance gain increases when we gradually decrease the token data size. 

\subsubsection{Experiment Set 3.2:}

We devised this experiment set in order to measure the effect of utilizing document and sentence data with really small data sizes similar to few-shot learning settings. Our results show significant improvement over baseline, and suggest that utilizing document and sentence data, both negative and positive, helps the model achieve a certain level of robustness.

\begin{figure}[!h]
\begin{center}
\includegraphics{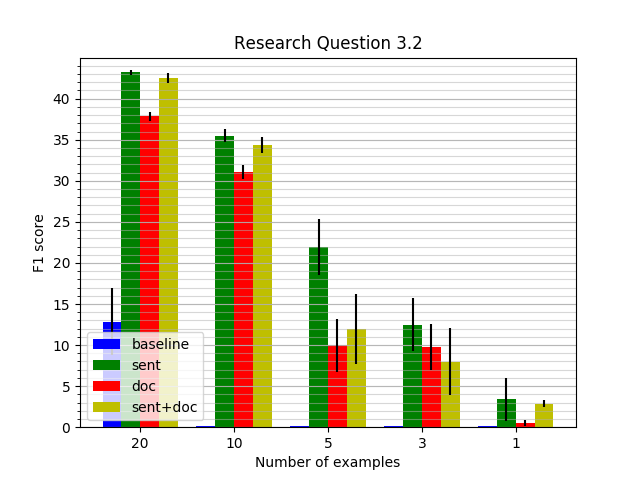} 

\caption{Results from experiment set 3.2, which is variation of experiment set 3 with tiny data sizes. Adding extra document or sentence labels with their respective classification tasks offers a great robustness to data size.}
\label{fig_rq3_tiny}
\end{center}
\end{figure}






\section{Experiment Details}
\label{expdetails}

\subsection{Training details}

The model processes a single document in each iteration. This is due to the fact that the encoder can process a maximum of $200$ sentences in each iteration given the GPU size. This is a huge bottleneck in terms of the speed of the training. To simulate mini-batching, which helps to stabilize the training process, we accumulate loss for $16$ iterations (documents). Each experiment is run for $30$ epochs and an intermediate score is calculated on a randomly sampled development set at the end of each epoch to select the best model. 

We use two instances of the Adam optimizer with weight decay~\cite{loshchilov2018decoupled}, one for each of the BERT encoder and the rest of the model since they have different learning rates. We use the recommended $2e^{-5}$ for the BERT encoder and $1e^{-4}$ for the rest of the model. We also utilize gradient clipping with a maximum norm of $1.0$.

\subsection{Model size}
We adhere to the original parameters and the size of the BERT encoder. The only difference is that we use $128$ instead of the original $512$ maximum tokens per input (sentence) due to the size issues it would create and distribution of sentence lengths in the data. BERT encoder outputs $128\times 768$ matrices for each sentence, a vector of $768$ length for each token.

The hidden size for both bi-GRU's is $512$, and their output for each input they receive is $1024$ since they are both bidirectional. The input size and hidden size for both the first (token) and second (sentence) MLP's are $1024$ and $4096$, respectively. For the final (document) MLP, input size and hidden size are $512$ and $2048$.

\chapter{Conclusion}
\label{conclusion}

We have investigated the effects of employing coarse-grained data for event information extraction. Our series of experiments show that training with sentence or document classification tasks help improve the results for token classification task. Our results indicate two main conclusions; first that adding documents that have no event information to the baseline offers an improvement, and second that employing coarse-grained data and tasks increase robustness as can be seen from the gradual decrease in performance compared to the rapid decline of the baseline with the change in data size.

\section{Future Work}
First and foremost, we plan on testing our hypothesis with different data sets and encoders. There are multiple event extraction data sets related to a variety of disciplines ranging from economics to biomedical research where we can implement our work. There are also questions concerning the effect of the pre-trained encoder we used. What would happen if we used a different Transformer-based encoder and its weights, especially an encoder pre-trained on sentences instead of documents? What about using random weights at the start of training or using a recurrent neural network based encoder?

We will also test the effects of utilizing more extra coarse-grained data. The other subtasks of the shared task we acquired our data are a possible source for the extra data. We utilized gold standard data throughout all our experiments. We will also be investigating the possible usage of silver coarse-grained data, which does not even require the considerable ease of labeling documents or sentences.

\bibliographystyle{apalike}
\bibliography{references}

\begin{thebibliography}{}

\bibitem[Caselli et~al., 2021]{caselli-etal-2021-protest}
Caselli, T., Mutlu, O., Basile, A., and H{\"u}rriyeto{\u{g}}lu, A. (2021).
\newblock {PROTEST}-{ER}: Retraining {BERT} for protest event extraction.
\newblock In {\em Proceedings of the 4th Workshop on Challenges and
  Applications of Automated Extraction of Socio-political Events from Text
  (CASE 2021)}, pages 12--19, Online. Association for Computational
  Linguistics.

\bibitem[Chen et~al., 2017]{chen2017automatically}
Chen, Y., Liu, S., Zhang, X., Liu, K., and Zhao, J. (2017).
\newblock Automatically labeled data generation for large scale event
  extraction.
\newblock In {\em Proceedings of the 55th Annual Meeting of the Association for
  Computational Linguistics (Volume 1: Long Papers)}, pages 409--419.

\bibitem[Chen and Qian, 2021]{chen2021description}
Chen, Z. and Qian, T. (2021).
\newblock Description and demonstration guided data augmentation for sequence
  tagging.
\newblock {\em World Wide Web}, pages 1--20.

\bibitem[Cho et~al., 2014]{cho2014properties}
Cho, K., Van~Merri{\"e}nboer, B., Bahdanau, D., and Bengio, Y. (2014).
\newblock On the properties of neural machine translation: Encoder-decoder
  approaches.
\newblock {\em arXiv preprint arXiv:1409.1259}.

\bibitem[Dale et~al., 2021]{dale-etal-2021-skoltechnlp}
Dale, D., Markov, I., Logacheva, V., Kozlova, O., Semenov, N., and Panchenko,
  A. (2021).
\newblock {S}koltech{NLP} at {S}em{E}val-2021 task 5: Leveraging sentence-level
  pre-training for toxic span detection.
\newblock In {\em Proceedings of the 15th International Workshop on Semantic
  Evaluation (SemEval-2021)}, pages 927--934, Online. Association for
  Computational Linguistics.

\bibitem[Devlin et~al., 2019]{Devlin+19}
Devlin, J., Chang, M.-W., Lee, K., and Toutanova, K. (2019).
\newblock {BERT}: Pre-training of deep bidirectional transformers for language
  understanding.
\newblock In {\em Proceedings of the 2019 Conference of the North {A}merican
  Chapter of the Association for Computational Linguistics: Human Language
  Technologies, Volume 1 (Long and Short Papers)}, pages 4171--4186,
  Minneapolis, Minnesota. Association for Computational Linguistics.

\bibitem[Dor et~al., 2019]{dor2019financial}
Dor, L.~E., Gera, A., Toledo-Ronen, O., Halfon, A., Sznajder, B., Dankin, L.,
  Bilu, Y., Katz, Y., and Slonim, N. (2019).
\newblock Financial event extraction using wikipedia-based weak supervision.
\newblock In {\em Proceedings of the Second Workshop on Economics and Natural
  Language Processing}, pages 10--15.

\bibitem[Elaraby and Litman, 2021]{elaraby-litman-2021-self}
Elaraby, M. and Litman, D. (2021).
\newblock Self-trained pretrained language models for evidence detection.
\newblock In {\em Proceedings of the 8th Workshop on Argument Mining}, pages
  142--147, Punta Cana, Dominican Republic. Association for Computational
  Linguistics.

\bibitem[Hu et~al., 2017]{meng-etal-cnn-iets-a-cnn}
Hu, M., Li, Z., Shen, Y., Liu, A., Liu, G., Zheng, K., and Zhao, L. (2017).
\newblock Cnn-iets: A cnn-based probabilistic approach for information
  extraction by text segmentation.
\newblock In {\em Proceedings of the 2017 ACM on Conference on Information and
  Knowledge Management}, CIKM '17, page 1159–1168, New York, NY, USA.
  Association for Computing Machinery.

\bibitem[Huang and Riloff, 2012]{huang2012bootstrapped}
Huang, R. and Riloff, E. (2012).
\newblock Bootstrapped training of event extraction classifiers.
\newblock In {\em Proceedings of the 13th Conference of the European Chapter of
  the Association for Computational Linguistics}, pages 286--295.

\bibitem[H{\"u}rriyeto{\u{g}}lu et~al., 2021]{hurriyetouglu2021multilingual}
H{\"u}rriyeto{\u{g}}lu, A., Mutlu, O., Y{\"o}r{\"u}k, E., Liza, F.~F., Kumar,
  R., and Ratan, S. (2021).
\newblock Multilingual protest news detection-shared task 1, case 2021.
\newblock In {\em Proceedings of the 4th Workshop on Challenges and
  Applications of Automated Extraction of Socio-political Events from Text
  (CASE 2021)}, pages 79--91.

\bibitem[Liu et~al., 2020]{Liu-et-al-2020-a-survey}
Liu, P., Wang, X., Xiang, C., and Meng, W. (2020).
\newblock A survey of text data augmentation.
\newblock In {\em 2020 International Conference on Computer Communication and
  Network Security (CCNS)}, pages 191--195.

\bibitem[Liu and Chen, 2015]{liu2015identifying}
Liu, X. and Chen, H. (2015).
\newblock Identifying adverse drug events from patient social media: a case
  study for diabetes.
\newblock {\em IEEE intelligent systems}, 30(3):44--51.

\bibitem[Loshchilov and Hutter, 2018]{loshchilov2018decoupled}
Loshchilov, I. and Hutter, F. (2018).
\newblock Decoupled weight decay regularization.
\newblock In {\em International Conference on Learning Representations}.

\bibitem[Lyu et~al., 2021]{lyu-etal-2021-zero}
Lyu, Q., Zhang, H., Sulem, E., and Roth, D. (2021).
\newblock Zero-shot event extraction via transfer learning: {C}hallenges and
  insights.
\newblock In {\em Proceedings of the 59th Annual Meeting of the Association for
  Computational Linguistics and the 11th International Joint Conference on
  Natural Language Processing (Volume 2: Short Papers)}, pages 322--332,
  Online. Association for Computational Linguistics.

\bibitem[Mintz et~al., 2009]{mintz-etal-2009-distant}
Mintz, M., Bills, S., Snow, R., and Jurafsky, D. (2009).
\newblock Distant supervision for relation extraction without labeled data.
\newblock In {\em Proceedings of the Joint Conference of the 47th Annual
  Meeting of the {ACL} and the 4th International Joint Conference on Natural
  Language Processing of the {AFNLP}}, pages 1003--1011, Suntec, Singapore.
  Association for Computational Linguistics.

\bibitem[Patra et~al., 2020]{patra-etal-2020-scopeit}
Patra, B., Suryanarayanan, V., Fufa, C., Bhattacharya, P., and Lee, C. (2020).
\newblock {S}cope{I}t: Scoping task relevant sentences in documents.
\newblock In {\em Proceedings of the 28th International Conference on
  Computational Linguistics: Industry Track}, pages 214--227, Online.
  International Committee on Computational Linguistics.

\bibitem[Peters et~al., 2018]{peters2018deep}
Peters, M.~E., Neumann, M., Iyyer, M., Gardner, M., Clark, C., Lee, K., and
  Zettlemoyer, L. (2018).
\newblock Deep contextualized word representations.

\bibitem[Ramshaw and Marcus, 1995]{ramshaw-marcus-95-text}
Ramshaw, L. and Marcus, M. (1995).
\newblock Text chunking using transformation-based learning.
\newblock In {\em Third Workshop on Very Large Corpora}.

\bibitem[Rei and S{\o}gaard, 2018]{rei-sogaard-2018-zero}
Rei, M. and S{\o}gaard, A. (2018).
\newblock Zero-shot sequence labeling: Transferring knowledge from sentences to
  tokens.
\newblock In {\em Proceedings of the 2018 Conference of the North {A}merican
  Chapter of the Association for Computational Linguistics: Human Language
  Technologies, Volume 1 (Long Papers)}, pages 293--302, New Orleans,
  Louisiana. Association for Computational Linguistics.

\bibitem[Shen et~al., 2020]{shen-etal-2020-hierarchical}
Shen, S., Qi, G., Li, Z., Bi, S., and Wang, L. (2020).
\newblock Hierarchical {C}hinese legal event extraction via pedal attention
  mechanism.
\newblock In {\em Proceedings of the 28th International Conference on
  Computational Linguistics}, pages 100--113, Barcelona, Spain (Online).
  International Committee on Computational Linguistics.

\bibitem[Vaswani et~al., 2017]{vaswani2017attention}
Vaswani, A., Shazeer, N., Parmar, N., Uszkoreit, J., Jones, L., Gomez, A.~N.,
  Kaiser, {\L}., and Polosukhin, I. (2017).
\newblock Attention is all you need.
\newblock In {\em Advances in neural information processing systems}, pages
  5998--6008.

\bibitem[Wang et~al., 2016]{wang-et-al-2016-a-multiple-instance}
Wang, W., Ning, Y., Rangwala, H., and Ramakrishnan, N. (2016).
\newblock A multiple instance learning framework for identifying key sentences
  and detecting events.
\newblock In {\em Proceedings of the 25th ACM International on Conference on
  Information and Knowledge Management}, CIKM '16, page 509–518, New York,
  NY, USA. Association for Computing Machinery.

\bibitem[Wei et~al., 2020]{wei2020study}
Wei, Q., Ji, Z., Li, Z., Du, J., Wang, J., Xu, J., Xiang, Y., Tiryaki, F., Wu,
  S., Zhang, Y., et~al. (2020).
\newblock A study of deep learning approaches for medication and adverse drug
  event extraction from clinical text.
\newblock {\em Journal of the American Medical Informatics Association},
  27(1):13--21.

\bibitem[Yarowsky, 1995]{yarowsky1995unsupervised}
Yarowsky, D. (1995).
\newblock Unsupervised word sense disambiguation rivaling supervised methods.
\newblock In {\em 33rd annual meeting of the association for computational
  linguistics}, pages 189--196.

\bibitem[Zhang et~al., ]{zhangcan}
Zhang, G., Lillis, D., and Nulty, P.
\newblock Can domain pre-training help interdisciplinary researchers from data
  annotation poverty? a case study of legal argument mining with bert-based
  transformers.

\bibitem[Zhao et~al., 2018]{zhao-etal-2018-document}
Zhao, Y., Jin, X., Wang, Y., and Cheng, X. (2018).
\newblock Document embedding enhanced event detection with hierarchical and
  supervised attention.
\newblock In {\em Proceedings of the 56th Annual Meeting of the Association for
  Computational Linguistics (Volume 2: Short Papers)}, pages 414--419,
  Melbourne, Australia. Association for Computational Linguistics.

\bibitem[Zhu et~al., 2015]{zhu2015aligning}
Zhu, Y., Kiros, R., Zemel, R., Salakhutdinov, R., Urtasun, R., Torralba, A.,
  and Fidler, S. (2015).
\newblock Aligning books and movies: Towards story-like visual explanations by
  watching movies and reading books.
\newblock In {\em Proceedings of the IEEE international conference on computer
  vision}, pages 19--27.

\end{thebibliography}

\appendix
\chapter{Detailed Results}

In this chapter of the appendices, tables with detailed results for all the experiments is listed. Each table contains a column named ``exp\_base" referring to the same baseline results for reference. ``sent", ``doc" and ``sent+doc" columns represent the usage of only sentence classification loss (variation 1), only document classification loss (variation 2), and both sentence and document classification loss (variation 3) in addition to token classification loss when training, respectively.

\begin{table}[ht]
\centering
\caption{Detailed results for experiment set 1, which focuses on the effects of our auxiliary tasks without any data addition.}
\resizebox{\textwidth}{!}{  
\begin{tabular}{|c|c|c|c|c|}
\hline
\textbf{\begin{tabular}[c]{@{}c@{}}\#num of \\ documents\end{tabular}} & \textbf{exp\_base} & \textbf{sent}     & \textbf{doc}     & \textbf{sent+doc}     \\ \hline
717                                                                    & 64.3420 $\pm$~0.2921  & 65.3298 $\pm$~0.1894 & 64.7353 $\pm$~0.2444 & 64.8228 $\pm$~0.3512 \\ \hline
645                                                                    & 64.3836 $\pm$~0.5550  & 64.0928 $\pm$~0.6175 & 64.5901 $\pm$~1.0282 & 64.1613 $\pm$~0.4205 \\ \hline
573                                                                    & 62.7110 $\pm$~0.5453  & 63.6060 $\pm$~0.0960 & 63.3323 $\pm$~0.7521 & 63.2426 $\pm$~0.3936 \\ \hline
501                                                                    & 62.1977 $\pm$~0.3972  & 61.9715 $\pm$~0.0423 & 62.7257 $\pm$~0.2095 & 62.9463 $\pm$~1.2105 \\ \hline
430                                                                    & 60.6711 $\pm$~0.5854  & 61.6818 $\pm$~1.0062 & 61.8722 $\pm$~0.1173 & 61.7387 $\pm$~0.8262 \\ \hline
358                                                                    & 61.0600 $\pm$~0.4680  & 60.8886 $\pm$~0.0812 & 60.2817 $\pm$~0.2523 & 60.8147 $\pm$~0.8071 \\ \hline
286                                                                    & 58.1234 $\pm$~0.6325  & 58.0093 $\pm$~0.6460 & 57.6096 $\pm$~0.1267 & 58.0629 $\pm$~0.4495 \\ \hline
215                                                                    & 55.4766 $\pm$~0.1252  & 54.4221 $\pm$~0.4860 & 55.2202 $\pm$~0.9379 & 56.1987 $\pm$~0.4816 \\ \hline
143                                                                    & 51.2678 $\pm$~0.5567  & 51.4187 $\pm$~0.4740 & 51.9452 $\pm$~0.4664 & 49.6417 $\pm$~1.0795 \\ \hline
71                                                                     & 39.2988 $\pm$~1.8238  & 39.5794 $\pm$~1.1141 & 41.0561 $\pm$~0.8125 & 40.1392 $\pm$~1.2548 \\ \hline
\end{tabular}
}
\label{rq1_table}
\end{table}

\begin{table}[ht]
\centering
\caption{Detailed results for experiment set 2, which focuses on the effect of adding negatively labeled documents with no event information.}
\resizebox{\textwidth}{!}{  
\begin{tabular}{|c|c|c|c|c|}
\hline
\textbf{\begin{tabular}[c]{@{}c@{}}\#num of\\ documents\end{tabular}} & \textbf{exp\_base} & \textbf{sent}     & \textbf{doc}     & \textbf{sent+doc}     \\ \hline
717                                                                   & 64.3420 $\pm$~0.2921  & 65.3298 $\pm$~0.1894 & 65.1503 $\pm$~0.4439 & 65.3221 $\pm$~0.2928 \\ \hline
645                                                                   & 64.3836 $\pm$~0.5550  & 64.0928 $\pm$~0.6175 & 64.0194 $\pm$~0.0571 & 65.0234 $\pm$~0.8096 \\ \hline
573                                                                   & 62.7110 $\pm$~0.5453  & 63.6060 $\pm$~0.0960 & 64.1297 $\pm$~0.7640 & 63.3397 $\pm$~0.5576 \\ \hline
501                                                                   & 62.1977 $\pm$~0.3972  & 61.9715 $\pm$~0.0423 & 63.0002 $\pm$~0.3124 & 62.7757 $\pm$~0.5070 \\ \hline
430                                                                   & 60.6711 $\pm$~0.5854  & 61.6818 $\pm$~1.0062 & 62.3406 $\pm$~0.0275 & 61.7372 $\pm$~0.3482 \\ \hline
358                                                                   & 61.0600 $\pm$~0.4680  & 60.8886 $\pm$~0.0812 & 61.8366 $\pm$~0.4351 & 61.4128 $\pm$~0.2562 \\ \hline
286                                                                   & 58.1234 $\pm$~0.6325  & 58.0093 $\pm$~0.6460 & 59.2479 $\pm$~0.7000 & 58.2156 $\pm$~0.9521 \\ \hline
215                                                                   & 55.4766 $\pm$~0.1252  & 54.4221 $\pm$~0.4860 & 55.9617 $\pm$~0.4265 & 56.1648 $\pm$~1.0320 \\ \hline
143                                                                   & 51.2678 $\pm$~0.5567  & 51.4187 $\pm$~0.4740 & 52.3773 $\pm$~0.7498 & 50.5637 $\pm$~3.4775 \\ \hline
71                                                                    & 39.2988 $\pm$~1.8238  & 39.5794 $\pm$~1.1141 & 43.2710 $\pm$~0.8947 & 43.9792 $\pm$~0.2619 \\ \hline
\end{tabular}
}
\label{rq2_table}
\end{table}

\begin{table}[ht]
\centering
\caption{Detailed results for experiment set 3, which focuses on the effects of adding extra coarse-grained data.}
\resizebox{\textwidth}{!}{  
\begin{tabular}{|c|c|c|c|c|c|}
\hline
\textbf{\begin{tabular}[c]{@{}c@{}}\#num of\\ token\\ annotated\\ documents\end{tabular}} & \textbf{\begin{tabular}[c]{@{}c@{}}\#num of\\ extra\\ auxiliary\\ data\end{tabular}} & \textbf{exp\_base} & \textbf{sent}     & \textbf{doc}     & \textbf{sent+doc}     \\ \hline
717                                                                                       & 0                                                                                    & 64.3420 $\pm$~0.2921  & 65.3298 $\pm$~0.1894 & 65.1503 $\pm$~0.4439 & 65.3221 $\pm$~0.2928 \\ \hline
645                                                                                       & 72                                                                                   & 64.3836 $\pm$~0.5550  & 64.5663 $\pm$~0.5121 & 64.2650 $\pm$~0.1842 & 65.2259 $\pm$~0.3991 \\ \hline
573                                                                                       & 144                                                                                  & 62.7110 $\pm$~0.5453  & 64.1659 $\pm$~0.8118 & 63.0057 $\pm$~0.4911 & 63.4717 $\pm$~0.0919 \\ \hline
501                                                                                       & 216                                                                                  & 62.1977 $\pm$~0.3972  & 63.4803 $\pm$~0.3426 & 62.7125 $\pm$~0.1686 & 63.4292 $\pm$~0.1548 \\ \hline
430                                                                                       & 287                                                                                  & 60.6711 $\pm$~0.5854  & 63.5683 $\pm$~0.4099 & 61.9322 $\pm$~0.9456 & 63.0599 $\pm$~0.0922 \\ \hline
358                                                                                       & 359                                                                                  & 61.0600 $\pm$~0.4680  & 61.7831 $\pm$~0.3405 & 61.4165 $\pm$~0.5701 & 61.5980 $\pm$~0.4963 \\ \hline
286                                                                                       & 431                                                                                  & 58.1234 $\pm$~0.6325  & 60.1496 $\pm$~0.2897 & 59.1478 $\pm$~0.1192 & 60.2890 $\pm$~0.2159 \\ \hline
215                                                                                       & 502                                                                                  & 55.4766 $\pm$~0.1252  & 57.7715 $\pm$~0.7474 & 56.2983 $\pm$~0.3384 & 58.2433 $\pm$~0.9674 \\ \hline
143                                                                                       & 574                                                                                  & 51.2678 $\pm$~0.5567  & 55.0612 $\pm$~0.6047 & 54.7686 $\pm$~0.6523 & 55.4178 $\pm$~0.1332 \\ \hline
71                                                                                        & 646                                                                                  & 39.2988 $\pm$~1.8238  & 49.4441 $\pm$~1.1924 & 49.7398 $\pm$~0.4377 & 51.8297 $\pm$~0.9912 \\ \hline
\end{tabular}
}
\label{rq3_table}
\end{table}

\begin{table}[ht]
\centering
\caption{Detailed results for experiment set 3.1, which is variation of experiment set 3 where extra data size is fixed.}
\resizebox{\textwidth}{!}{  
\begin{tabular}{|c|c|c|c|c|c|}
\hline
\textbf{\begin{tabular}[c]{@{}c@{}}\#num of\\ token\\ annotated\\ documents\end{tabular}} & \textbf{\begin{tabular}[c]{@{}c@{}}\#num of\\ extra\\ auxiliary\\ data\end{tabular}} & \textbf{exp\_base} & \textbf{sent}     & \textbf{doc}     & \textbf{sent+doc}     \\ \hline
358                                                                                       & 359                                                                                  & 61.0600 $\pm$~0.4680  & 61.7831 $\pm$~0.3405 & 61.4165 $\pm$~0.5701 & 61.5980 $\pm$~0.4963 \\ \hline
286                                                                                       & 359                                                                                  & 58.1234 $\pm$~0.6325  & 59.8364 $\pm$~0.2759 & 59.7761 $\pm$~0.8661 & 59.7066 $\pm$~0.2504 \\ \hline
215                                                                                       & 359                                                                                  & 55.4766 $\pm$~0.1252  & 57.8553 $\pm$~0.5770 & 56.6985 $\pm$~0.5273 & 58.2742 $\pm$~0.9637 \\ \hline
143                                                                                       & 359                                                                                  & 51.2678 $\pm$~0.5567  & 54.6631 $\pm$~0.5420 & 53.9005 $\pm$~0.9373 & 54.7954 $\pm$~0.4511 \\ \hline
71                                                                                        & 359                                                                                  & 39.2988 $\pm$~1.8238  & 49.0730 $\pm$~0.3271 & 48.7882 $\pm$~0.2006 & 49.6189 $\pm$~0.7479 \\ \hline
\end{tabular}
}
\label{rq3_fix_extra_table}
\end{table}


\begin{table}[!ht]
\centering
\caption{Detailed results for experiment set 3.2, which is variation of experiment set 3 with tiny data sizes.}
\resizebox{\textwidth}{!}{  
\begin{tabular}{|c|c|l|c|c|}
\hline
\textbf{\begin{tabular}[c]{@{}c@{}}\#num of \\ token\\ annotated\\ documents\end{tabular}} & \textbf{exp\_base} & \multicolumn{1}{c|}{\textbf{sent}} & \textbf{doc}      & \textbf{sent+doc} \\ \hline
20                                                                     & 12.8237 $\pm$~4.0776  & 43.2263 $\pm$~0.3178                  & 37.8599 $\pm$~0.5719 & 42.5216 $\pm$~0.5838 \\ \hline
10                                                                     & 0.0000 $\pm$~0.0000   & 35.4900 $\pm$~0.7729                  & 31.1247 $\pm$~0.8497 & 34.3691 $\pm$~1.0103 \\ \hline
5                                                                      & 0.0000 $\pm$~0.0000   & 21.9365 $\pm$~3.4075                  & 9.9743 $\pm$~3.1956  & 11.9494 $\pm$~4.2202 \\ \hline
3                                                                      & 0.0000 $\pm$~0.0000   & 12.4881 $\pm$~3.2203                  & 9.7442 $\pm$~2.7849  & 7.9693 $\pm$~4.0605  \\ \hline
1                                                                      & 0.0000 $\pm$~0.0000   & 3.3871 $\pm$~2.5824                   & 0.5398 $\pm$~0.3380  & 2.8559 $\pm$~0.4523  \\ \hline
\end{tabular}
}
\label{rq3_tiny_table}
\end{table}

\end{document}